# Fully Automated Binary Pattern Extraction For Finger Vein Identification using Double Optimization Stages-Based Unsupervised Learning Approach

Ali Salah Hameed[1], Adil Al-Azzawi[2]
Computer Science Department
College of Science, University of Diyala
Diyala, Baqubah, Iraq
scicompms2118@uodiyala.edu.iq[1], adil_alazzawi@uodiyala.edu.iq[2]

*Abstract*—Today, finger vein identification is gaining popularity as a potential biometric identification framework solution. Machine learning-based unsupervised, supervised, and deep learning algorithms have had a significant influence on finger vein detection and recognition at the moment. Deep learning, on the other hand, necessitates a large number of training datasets that must be manually produced and labeled. In this research, we offer a completely automated unsupervised learning strategy for training dataset creation. Our method is intended to extract and build a decent binary mask training dataset completely automated. In this technique, two optimization steps are devised and employed. The initial stage of optimization is to create a completely automated unsupervised image clustering based on finger vein image localization. Worldwide finger vein pattern orientation estimation is employed in the second optimization to optimize the retrieved finger vein lines. Finally, the proposed system achieves 99.6 - percent pattern extraction accuracy, which is significantly higher than other common unsupervised learning methods like k-means and Fuzzy C-Means (FCM).

## I. INTRODUCTION

Biometric technology has gotten a lot of attention from the public in recent years in which a remarkable increasing need for biometric system safety and accuracy was observed [1]. Fingerprints, palm prints, iris, gait, facial characteristics, voice recognition, signature, heartbeat, and palm vein have all been suggested as biometric identification methods. However, the majority of these biometric identification methods have a number of flaws, including susceptibility, lighting, facial expression, position, and occlusion. Instead, a more reliable, safe, and convenient biometric technique, the finger vein, has lately been deployed [2].

When it comes to finger vein recognition, convolutional neural networks (CNNs) have become increasingly popular in the field of image recognition as deep learning technology has advanced and the pixel-level label information obtained by traditional texture extraction techniques has been used to train CNNs [3]. A template matching strategy is used to classify the images of finger veins based on the feature data. Finger vein patterns were extracted using CNN and then restored using Fully Convolutional Network (FCN) according to Qin et al.'s [4].

However, designing a completely automated finger-vein detection system in the absence of labeled data continues to be a difficult undertaking to do. Aside from that, in the most recent suggested systems, the entire data set (the entire finger vein image) has been utilized in order to train the model, which results in some biasing in terms of the data and environmental similarities, such as skin color, backdrop, and other aspects. On the other hand, the finger vein lines should be regarded as the primary pure data because they are the primary pattern upon which it is built. This work proposes a completely automated, unsupervised learning strategy for the automatic production of training finger vein line data. Our technique eliminates the challenges associated with manually labeling training datasets by employing a fully automated way to build them based on certain complex algorithms and procedures [5].

## II. RELATED WORKS

Finger vein recognition systems were of interest to several researchers. This is a list of some of the most recent publications that are related to the subject of this paper [6] [7].

DSP platform-enabled finger vein-recognition system was developed by Zhi Liu et al. [8] can authorize an individual in around 0.08 seconds and has an Accuracy of 0.07 % on a 100-subject dataset. Ajay Kumar et al. [9] developed a fully automated and extensive method for comparing finger sample images that use the subsoil properties of the finger, i.e., finger texturing and finger-vein images. It can derive feature information through a variety of approaches, such as the Gabor filter, matching filter, repeating line tracing, and maximal curve extraction. Zhi Liu et al. [10] Wonseok Song et al. introduced a mean curvature-based finger vein verification method that treats the vein picture as a geo-metric form and discovers valley-like structures with negative mean curvatures. The matching of vein patterns using matched pixel ratio demonstrates that, while keeping minimal complexity, the suggested technique achieves 0.25 percent EER through veins appearing as dark shadow lines in the acquired picture as hemoglobin occurs densely in blood vessels [8] [9] [10].

## III. BACKGROUND THEORY

Unsupervised learning relies heavily on the idea of clustering. The main goal of this research is to find patterns and structures in an unstructured dataset. Clustering or grouping dataset that are most similar with each other is the major objective. For instance, some clustering adopts the distance majority method to select the proper cluster for each data point, just like any other data point without the label. K-means clustering is clarified in Algorithm (1) shown below [11] [12].

**Algorithm (1): Unsupervised Clustering Algorithm**

1) Cluster **centroids** should be set up. $\mu_1, \mu_1, \ldots, \mu_k \in \mathbb{R}^n$ With $K=5$, where $K$ is the number of clusters.

2) Repeating

3) For **each** single pixels $P^{(i)}$ do (1)
$$P^i = \min_j \|x^{(i)} - \mu_j\|^2 \quad (1)$$

4) On a per-cluster basis $(j)$ do (2)
$$\mu_j = \frac{\sum_{i=1}^n 1\{P^{(1)}=1\}x^{(i)}}{\sum_{i=1}^n 1\{P^{(1)}=j\}} \quad (2)$$

5) Loop $(j)$ is completed // [end of for].

6) Loop $(i)$ is completed // [end of for].

7) End

It is also possible to utilize $K$-Means to locate a potentially harmful region in a photograph. Cluster mean is used to identify a set of clusters $(P_j)$ [11] [12].

## IV. THE METHODOLOGY AND SYSTEM DESIGN

The methodology of the proposed approach includes two steps of optimization for a completely automated finger vein pattern extraction based on unsupervised learning, as shown in Fig.1. The initial level of optimization involves creating a completely automated, unsupervised finger vein image localization utilizing an efficient image clustering technique. The second optimization stage involves creating a Global Pattern Optimization (GPO) based on the extraction, indication, and optimization of finger vein lines.

The suggested system is divided into three stages, the first of which is per-processing. At this stage, certain preliminary image processing procedures are used to increase the image's contrast and mid-range intensity stretching (optimization). The second level is completely automatic, unsupervised image localization. At this level, a completely automated unsupervised image clustering technique is used to overcome the major shortcomings of typical unsupervised learning approaches (such as the $K$-means). Our solution covers the difficulties of clustering random initialization, cluster assignment, and time consumption. In the final step, global finger vein lines are extracted using global pattern optimization. The GPO projection is used to correct the finger vein lines that have been recovered.

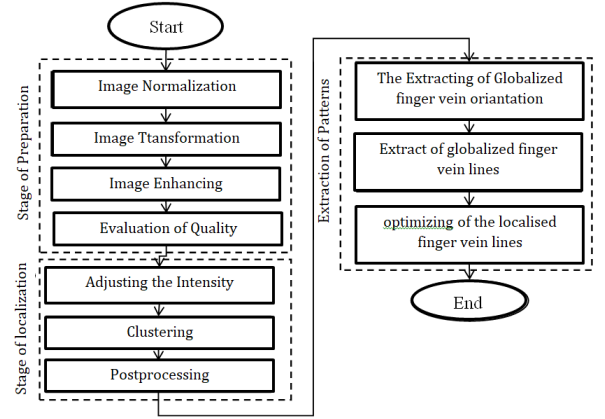

Fig. 1. Fully automated Finger vein pattern extraction based on double optmization stages.

### A. Pre-processing stage (Stage of preparation):

The preprocessing stage (Stage of preparation) major procedures revolve around adjusting the image intensity level. The basic notion behind this is that defining the number of desirable clusters is the most challenging stage in building a fully automated unsupervised learning technique. Clustering stabilization, on the other hand, is a challenge that our design considered and resolved. Concerning the first issue, identifying the relevant number of clusters, we projected the original intensity level, which is normally between [0] and [255] in the majority of cases. Because the image intensity has been standardized to be between [0] and [1], we will only have 10 differences. Then, we employed mid-range intensity stretching to reduce the intensity to five levels rather than ten. In this scenario, we are very certain that there are only 5 total clusters in the domain which is utilized to compute the cluster number. Applying pre-processing depends on the following steps:

1. *Preparing Normalized Intensity (Image normalization):* The image normalization procedure is carried out using a technique called localized image normalization, which is based on estimating the local mean and variance utilizing the intensity of the vein in the middle of the fingers. Equation (3) depicts the fundamental idea of localized image normalizing of finger vein images [13].

$$x'(i,j) = \sqrt{\frac{\sigma_l(\mathcal{I}(i,j) - \bar{x}_l)^2}{\sigma}} \quad (3)$$

Where $\bar{x}$ is the average of pixel intensities and $\sigma$ is used to calculate the standard deviation. [13].

2. *Image Smoothing and Noise Removal for Finger Veins (Image Enhancing):* The technique of modifying an image such that the output is better suited than the original is known as a finger vein image enhancement. In this work, we employ the filter called Wiener used for smoothing and removing noise from finger vein images. The inverse filtering and noise smoothing procedures effectively reduce a total mean square error. Equation (4) denotes the Wiener filter numerical representation [9] [10].

$$Weiner(f_1, f_2) = \frac{H(f_1, f_2) \times Sxx(f_1, f_2)}{|H(f_1, f_2)|^2 \, Sxx(f_1, f_2) + S\eta\eta(f_1, f_2)} \quad (4)$$

For the original image and for the added noise, the power spectrums are $Sxx(f_1, f_2)$ and $S\eta\eta \ (f_1, f_2)$ respectively. $H(f_1, f_2)$ Is the blurring filter. Fig.2 (c) shows the result of applying Wiener filtering to the noise reduction process [14] [15].

3. *Image Adjustment by mid-range stretching:* Image correction is a popular pre-processing technique for improving images. Equation (5) is the fundamental image adjustment equation [16].

$$X_{ij} = \frac{L_{out} + (H_{out} - L_{out}) \times (X_{ij} - L_{in})}{(H_{in} - L_{in})} \quad (5)$$

There are two parameters in the output image: the lower and higher bounds of intensity level, $L_{out}$ and $H_{out}$. $L_{in}$ And $H_{in}$ represent the input images lowest and greatest intensity levels, respectively. The major objective of this step is to execute image intensity level quantification after normalization the finger vein image. If we use the image after it has been adjusted, then the intensity levels will range from [0] to [1], with [0] referring to black and [1] referring to white. Starting with the premise of equalizing image intensity levels to be the initial cluster number to start with, appropriate cluster size can be determined. In our dataset, we have an adjustment factor of [0.2-0.6]. For example, where 0.2 signifies the lowest intensity level and 0.6 indicates the highest. As we can see, there are five intensity levels ranging from 0.2 to 0.6, so we'll use that as our beginning cluster number. Fig .3 (b) Illustrates the outcome.

*B. Finger Vein Localizing Stage (Stage of localization):*

There have been numerous kinds of clustering algorithms developed, and they may be summarized as follows: Methods of partitioning, hierarchical, model, density-based, and grid-based approaches. Traditional clustering methods such as $K$-means and FCM are employed, as well as our own intensity-based clustering methodology. An intensity distribution model is used in the clustering method, $P(i \ ; \ d)$, which describes the relationship between the intensity difference value $d$ and signed difference intensity values. For instance, let $\{I_1, I_2 \ldots I_n\}$ represents a collection of images taken with the same modality and anatomical structure on different people, while $\{x^{(1)}, x^{(2)} \ldots x^{(L)}\}$ denotes the number of "clusters" formed from all $L$ pixels in an image, with the number determined by the size of the intensity interval between the pixels. There are four initial cluster levels if the adjusted intensity range is set between [0.2, 0.8] with an increment value of 0.15. Therefore, the cluster ranges are going to [0.2-0.35], [0.35-0.5], [0.5-0.65], and [0.65-0.8] for Cluster 1, Cluster 2, Cluster 3, and Cluster 4 respectively. In this case, $x^{(i)}$ is a real intensity value in a specified range, where $1 <= i <= L$. Let $\{\theta_1, \theta_2 \ldots \theta_K\}$ be the set of the average intensity values of $K$ clusters. Let $U_j$ be the index of the cluster whose center $(\theta_j)$ is closest to $x^{(i)}$. The cluster assignment of all pixels $\{\langle x^{(1)}, U_j \rangle, \langle x^{(2)}, U_2 \rangle, \ldots, \langle x^{(n)}, U_n \rangle\}$ is repeatedly updated based on the average intensity $(\theta_j)$ of clusters.

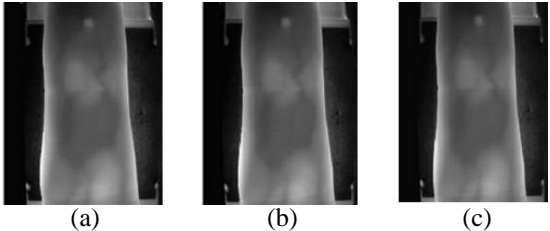

Fig. 2. Results of the preprocessing stage for the finger vein (a) original finger image, (b) outcome of the image normalization, (c) the outcome of image enhancement and noise reduction.

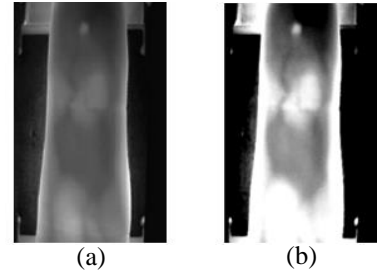

Fig. 3. Using Image Adjustment, a mid-range stretch is achieved (a) original image, (b) corrected (adjusted) image.

As stated in (6), the centers $(\theta_j)$ of the $K$ clusters are established as equally dispersed intervals across the intensity range with an equal step size as given below:

$$Step_{size} = \frac{I_{Range}}{K \times 0.1} \times 0.1 \quad (6)$$

Where $I_{Range}$ denotes to the contrast between the two images (greatest and lowest intensity levels).

In Algorithm (2), the proposed optimization method is illustrated. Instead of using the standard [0-255] band of 256 intensity levels, this method uses the [0-1] range from 10 intensity levels, subsequently reducing the number of degrees to 5.

**Algorithm (2): Unsupervised Clustering Optimization**

1. Convert Image from $2D$ to $1D_{length}$
$$Length = row \times col \quad (7)$$
2. While configuring the parameters, use the previously chosen intensity images modification level to establish the cluster count.
3. Create vectors for the initialized cluster.
$$Inatialized = (1D_{Cluster \, id})_{1 \times length} \quad (8)$$
4. The initial value of the cluster centers may be determined by using the equations below :
$$Max_{value} = \max imum(gray\_level) \quad (9)$$
$$Min_{value} = \minimum(gray\_level) \quad (10)$$
$$Span = Max_{value} - Min_{value} \quad (11)$$
$$Stair_v = \frac{Span}{Cluster_{id}} \quad (12)$$

5. $cluster_{centriods}$  $\mu_1, \mu_1, \ldots, \mu_k \in \mathbb{R}^n$ must be Initialized using(13).
$$increment_v = Stair_v \quad (13)$$
6. Begin For $i=1$ to $Cluster_{id}$ do
 Set up the cluster centers depending on:
$$Cluster_c(i) = increment_v \quad (14)$$
$$increment_i(i) = Span_v + increment_v(i-1) \quad (15)$$
7. The starting value of the means must be supplied.
$$mean\,(i) = 2 \quad (16)$$
8. Assemble $Update_i = 0$.
9. Loop $(i)$ is completed // [end of for].
10. Repeating
11. Begin For loop $i$ from the start to the finish (image length) then you must do
12. Begin For $j$ loop with relation clustered id then you must do
$$temp \leftarrow 1Dim[i] \quad (17)$$
$$dif^{(i)} := abs \left\| temp^{(i)} - Cluster_{center_j} \right\| \quad (18)$$
13. Loop $(j)$ is completed // [end of for].
$$y^{(i)} := \arg\min_j \left\| dif^{(i)} \right\| \quad (19)$$
14. If the tested pixel is the lowest Index $y^{(i)}$ Inside the Cluster then this pixel should be assigned to the cluster.
15. Ending if
16. Loop $(i)$ is completed // [end of for].
17. Determine how many of each type have been evaluated $Cluster_{Id}$.
18. Updating, the mean values
$$Cluster_{c_j} = \mu_j \quad (20)$$
$$\mu_j := \frac{\sum_{i=1}^k 1\{cluster^{(1)}=i\}}{\sum_{i=1}^k 1\{count^{(1)}=j\}} \quad (21)$$
19. Updates will continue until the coverage is complete.
20. End

C. *Pattren Extraction of Finger Vein(Extraction of Pattern ):* Two stages are presented and implemented in this stage. The global pattern of the finger vein line was calculated in the first step, and the globalized finger vein pattern orientation was estimated and retrieved in the second step. Finally, the globalized pattern orientation and frequency estimates are used to optimize and then draw the specific vein patterns on the fingers. Applying Finger Vein Pattren Extraction depends on the following steps:

1. *Estimation of the Orientation of Finger Vein Patterns on a Global Scale:* After normalizing the images in the previous step, we can now go to the next step. This step involves calculating the direction of the specific finger vein characteristics. Finger vein images are represented by the orientation image, which reflects their properties. Ridges and furrows invariant coordinates are specified by the localized feature orientation. Algorithm (3) displays the major phases of the technique of estimating the feature orientation.

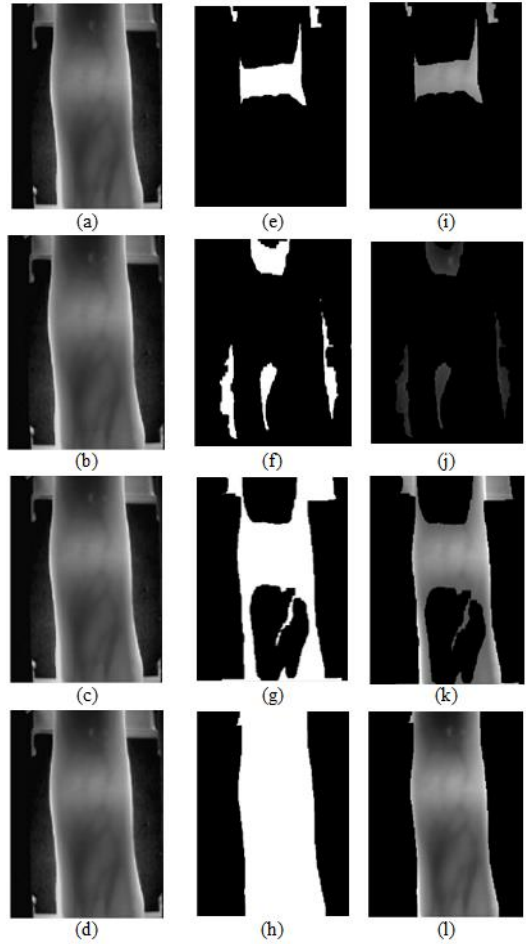

Fig. 4. On the basis of many universal clustering techniques, Finger Vein Localization (a), (b), (c), and (d) are the orginal image, (e), (f), (g), and (h) are the k-means, FCM, Otsu's double Thresholding, and our proposed clustering, respectively, are the clustering results. (i), (j), (k), and (l) There are localization findings for k-means , FCM Otsu's double Thresholding as well as our proposed algortihm .

2. *Globalized Finger Vein Pattern Frequency Estimation*: The normalized or adjusted image is utilized to calculate the feature patterns frequencies of the image of a finger vein, which is dependent based on the predicted direction of the isolated feature. Vein lines can be detected in this scenario as an indication shape (waveform) oriented in the usual direction of the local ridge direction. Assume that $\mathcal{G}$ represents the normalized finger vein images and $\mathcal{O}$ represents the local ridge (vein lines) orientation produced in the previous phase. Then, to estimate the frequency of the localized image, we  need use Algorithm (4).
3. *Extraction of Finger Vein Lines from a Specific Area*: The second optimization step, which is the basis for the finger vein pattern filtration model, is called the optimization phase. Algorithm (5) describes how to connect and optimize a finger vein line at a certain location.

**Algorithm (3): Globalized Pattern Orientation**

**Inputs:** $\mathcal{G}$: Finger vein images that have been normalized.

**Output:** $\mathcal{O}$: The position of the local ridges.

1. Images were divided into $w \times w$ blocks from the normalized image $\mathcal{G}$
2. Determine the image variations that employ the Sopel Method in different positions, $x - axis$ $Gradx(i,j)$ and $y - axis$ $Grady(i,j)$ for every pixels in the normalized image. The following formula is used to determine the variations magnitude:
$$|G(i,j)| = \sqrt{Grad_x(i,j)^2 + Grad_y(i,j)^2} \quad (22)$$
3. The estimated magnitude was determined by:
$$|G| = |Grad_x(i,j)| + |Grad_y(i,j)| \quad (23)$$
4. Using the $point\ (i,j)$ as a reference point, calculate the localized orientation of each single block.
$$Ve_x(i,j) = \sum_{u=i-\frac{w}{2}}^{i+\frac{w}{2}} \sum_{i-\frac{w}{2}}^{j+\frac{w}{2}} (2Grad_x(m,n)2Grad_y(m,n)) \quad (24)$$
$$Ve_y(i,j) = \sum_{u=i-\frac{w}{2}}^{i+\frac{w}{2}} \sum_{i-\frac{w}{2}}^{j+\frac{w}{2}} \left(Grad_x^2(m,n) - Grad_y^2(m,n)\right) \quad (25)$$
$$Angle(i,j) = \frac{1}{2} tan^{-1}\left(\frac{Ve_y(i,j)}{Ve_x(i,j)}\right) \quad (26)$$

Where Angle(j, j) the direction of each block's local ridge is represented by the Least Square Estimation.

5. Use the following formula to turn the 2D image into a continuous vector field:
$$\euro_x(a,b) = \cos(2Angle(i,j)) \quad (27)$$
$$\euro_y(a,b) = \sin(2Angle(i,j)) \quad (28)$$
6. Compute the direction of localized ridges for each $point(x,y)$ using:
$$\mathcal{O}(x,y) = \frac{1}{2} tan\left(\frac{\euro_x'(a,b)}{\euro_y'(a,b)}\right) \quad (29)$$
7. End

**Algorithm (4): Globalized Pattern Frequency Estimation**

Inputs: $\mathcal{G}$: The position of the local ridges.

Output: $\Omega$: The assessment of the frequency of ridges.

1. The normalized image $\mathcal{G}$ was divided into a series of blocks with the dimensions $w \times w$ (16 × 16).
2. For every block which is located in the center of the $pixel(i,j)$ do
3. Calculate the size of the locally oriented window $l \times w$, making use of window dimensions $(32 \times 16)$.
4. For every block which is located in the center of the $pixel(i,j)$ do
5. Create a 1D vector and use it to compute the x-signature values.
$$X[k] = \frac{1}{w}\sum_{d=0}^{w-1} \mathcal{G}(u,v), k = 0,1,\dots,l-1 \quad (30)$$
$$U = i + \left(d - \frac{w}{2}\right)cos\mathcal{O}(i,j) + \left(k - \frac{l}{2}\right)sin\mathcal{O}(i,j) \quad (31)$$
$$V = j + \left(d - \frac{w}{2}\right)cos\mathcal{O}(i,j) + \left(k - \frac{l}{2}\right)sin\mathcal{O}(i,j) \quad (32)$$
6. Ending nested loop / [end of for].
7. Ending main loop / [end of for].

8. Estimate the frequency ridge from the 1D vector of the x-signature by allowing $\mathcal{T}(i,j)$ be the averaged pixels in the x-signature between two peaks. After that, the frequency $\Omega(i,j)$ calculated using (33).
$$\Omega(i,j) = \frac{1}{\mathcal{T}(i,j)} \quad (33)$$
9. End

**Algorithm (5) Localized Finger Vein Pattern Extraction**

**Inputs:**
  $Img$ : Image of a Finger Vein that has been localized
  $mask_{hight}$ : The mask's overall height
  $mask_{width}$: The mask's overall width

**Output:**
  $region$: The finger vein area is shown with a binary mask.

1. Extraction of the image size is required for the initialization.
2. By inspecting the bottom half of the supplied finger vein image, Equation (34) can be utilized to establish the beginning location.
$$mask = Zeros(mask_{hight}, mask_{width}) \quad (34)$$
3. Create the pretreatment filtering mask for use in post-processing.
$$mask = \sum_{i=1}^{hight/2} mask = -1 \quad (35)$$
$$mask = \sum_{i=\frac{hight}{2}}^{n} mask = 1 \quad (36)$$
4. By adjusting the mask size x and y, you may create the kernels of the filter depending on the value of sigma you've chosen.
$$kernel = \frac{1}{2 \times \pi \times \sigma} e^{-\frac{x^2+y^2}{2 \times \sigma^2}} \quad (37)$$
5. Carry out the actual image filtering operations
$$Image_{filtered} = Conv(image, kernel) \quad (38)$$
6. Use the greatest curvature technique to determine image curvatures.
7. Extricate the $x, y, z$ coordinates from the coordinate system.
8. Track and find the lines in the image of the veins on the finger.
9. For each block size $n \times n$ check the global orientation-based Algorithm (3).
10. Use the line srt to close the image from the $n \times n$ block that is extracted in the previous step and with $\theta$ degree that is also extracted from the block size in Algorithm (3).
11. The binary image should be cleaned up of any extraneous pixels.
12. End

Fig. 5. (a), (b), and (c) represents the outcomes of the globalized finger vein estimate stage for the localized finger feature extraction.

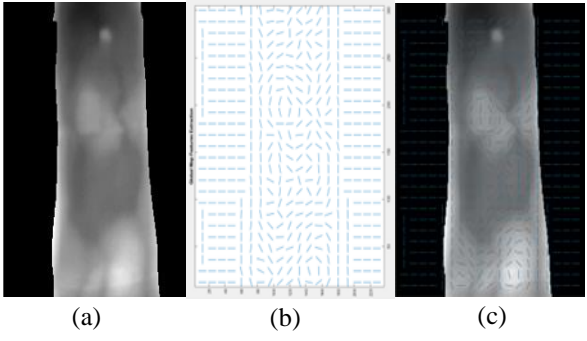

Fig.5. Globalized Pattren Orientation Estimation Result (1) The Original image, (3) Globalized image feature orientation estimation.

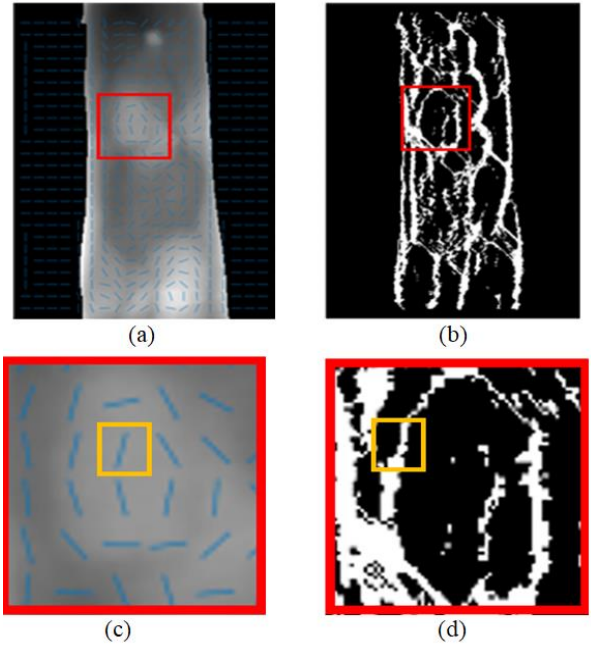

Fig.6. Localized image Pattern Extraction results (a) Globalized image feature orientation estimation, (b) Finger vein line extraction based curvatures model, (c) One block of the orientation estimation, (d) Image closing using line str from the n × n block that is extracted in the previous step and with θ degree.

Fig.6. depicts the final optimization results, which are based on the GPO Estimation algorithm. For each block, we employed the same structural line and degree. Once the small object pixels in the digital finger vein line image have been eliminated; they are removed using the binary image closure approach.

## V. EXPERMENTAL RESULTS

### A. Dataset

The finger vein database SDUMLA-HMT was used in this investigation. It was created by Wuhan University's Joint Lab for Smart Technology and Machine Intelligence. The database includes multiple images with three type index, middle and ring for left and right hands of an individual. Each of these fingers has its data collected six times, resulting in six images of the finger veins [17].

### B. Evaluation Criteria

Preprocessing, clustering, and finger vein, binary line extraction play a role in the performance evaluation of our suggested system. Our proposed system preprocessing architecture is initially evaluated utilizing the most often employed image preprocessing measures, such as peak signal to noise ratio (P-S-N-R) (39), mean square error (M-S-E) (40) and signal to noise ratio (S-N-R) (41), [18] [19][20][21][22][23].

$$PSNR = 10 \times log_{10}\left(\frac{MAX_I^2}{MSE}\right) \quad (39)$$

Here, $MAX_I$ is the image's highest potential pixel value, while $MSE$ is the image's average squared error.

$$MSE = \frac{1}{n \times m}\sum_{i=0}^{n-1}\sum_{j=0}^{m-1}[O(i,j) - E(i,j)]^2 \quad (40)$$

Where $(O(i,j))$ is the observed value, and $(E(i,j))$ is the predicted value.

$$SNR = 10 \times log_{10}\left(\frac{P_{signal}}{P_{noise}}\right) \quad (41)$$

$P_{signal}$ Is the image's average value, while $P_{noise}$ is the normal image division.

We also outperform current clustering methods including k-means, FCM, Otsu's local threshold, and the approach. The time required to perform each clustering algorithm is calculated. A number of factors are taken into consideration when determining how well the finger vein pattern extraction performs:

1. Accuracy (Recognition Rate): How many correct recognition judgments are made compared to the total number of tries? As expressed in (42) [18] [19][20][21][22][23].

$$Accuracy = \frac{TP}{TP+TN} * 100 \quad (42)$$

2. Recall (Sensitivity): The number of relevant predictions divided by the number of relevant findings, As provided in (43) [18] [19][20][21][22][23].

$$Recall = \frac{TP}{TP+TN} \quad (43)$$

3. Precision: Negative detection rate is defined as the percentage of negative predictions that are really correct. As provided in (44) [18][19][20][21][22][23].

$$Precision = \frac{TP}{TP+FN} \quad (44)$$

4. F1-Measurement: it's a harmonic mean between precision and sensitivity, As seen in (45) [18] [19][20][21][22][23].

$$F1 - Measurement = 2 \times \frac{2TP}{2TP+FP+FN} \quad (45)$$

### C. Performance Results

There are measures for pre-processing images such as: The ratio of signal to noise, Mean or Average Square error and Signal-to-Noise Ratio at the Peak or Maximum that are shown in TABLE I.

TABLE I. PREPROCESSING RESULT FOR ORGINAL AND ENHANCED IMAGES.

| Original Images Result | | | Preprocessed Images Result | | |
|---|---|---|---|---|---|
| *P- S-N-R* | *M-S-E* | *S-N-R* | *P- S-N-R* | *M-S-E* | *S-N-R* |
| 63.3667 | 0.05505 | 0.91529 | 66.9896 | 0.01978 | 0.96670 |

Using an unsupervised clustering approach, the average time for the finger vein image localization step is presented in Table II. By an average of 77.53 seconds, our optimization method is significantly quicker than traditional k-means and the FCM.

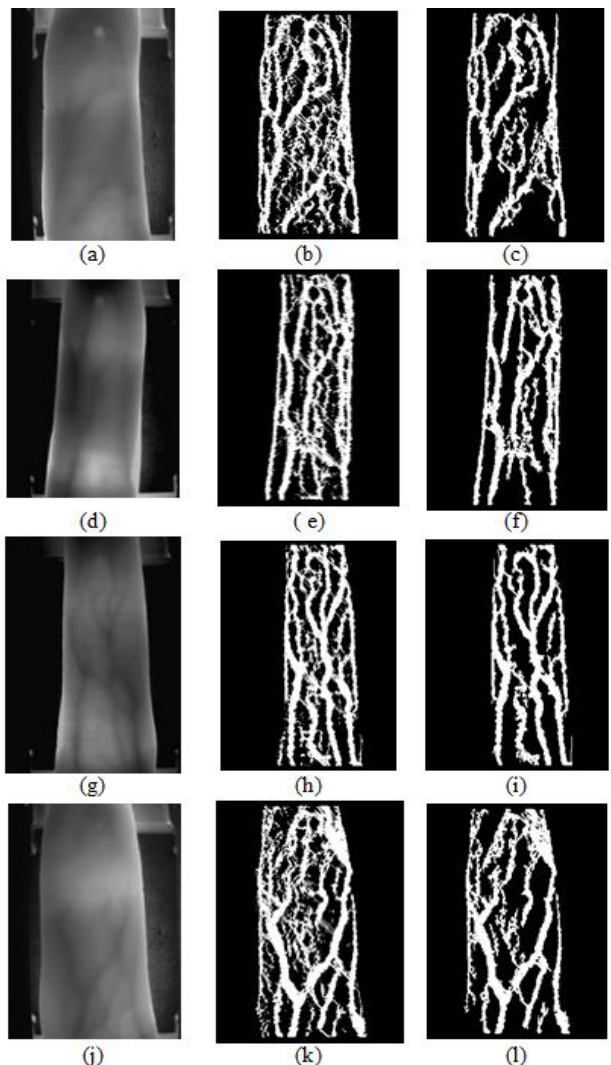

Fig.7. Outcomes of localized image optimization using several universised procedure utilized (a), (d), (g), and (j) the orginal image, (b), (e), (h), and (k) the finger vein line extraction based on repeated line tracking method. (c), (f), (i), and (l) the image line extraction based on (GPO).

TABLE II. TIME SPENT ON AVERAGE CLUSTERING IMAGES OF FINGER VEINS

| Algorithm name | Spending time |
|---|---|
| Traditional K-means | 16.01820 s |
| Fuzzy C-Means Clustering | 2.83290 s |
| Otsu's Threshold | 0.00156 s |
| **Our Optimization Clustering** | 0.82967 s |

Rate of Recognition, Precision, Sensitivities and Specificity are used to evaluate the performance of the localization and grouping stage. The formula for calculating these measurements are given in (42), (43), (44), and (45), and Our proposed Performance Results are shown in Table III.

TABLE III. THE IMPLICATIONS OF OUR PROPOSED ACTIVITIES

| Algorithm name | Accuracy | Precession | Recall | F1-measue |
|---|---|---|---|---|
| k-mean. | 22.67913 | 77.33376 | 19.90343 | 30.4792 |
| Fuzzy C-Means. | 19.97815 | 66.36706 | 21.02692 | 26.7769 |
| Otsu's Threshold. | 76.52724 | 91.60966 | 46.11347 | 61.14073 |
| **Our approach.** | **99.56007** | **90.65569** | **55.44011** | **67.74641** |

Table III shows that our suggested method (Finger vein binary pattern extraction) achieved a better accuracy (99.7 percent) compared to the other unsupervised clustering algorithms. On average, our optimized finger vein identification based on binary pattern extraction has improved the accuracy by 23% compared with the second heist accuracy that has been achieved by the k-means. Moreover, comparing with the other metrics, our approach has achieved better precession, Recall, and F1-measue by (91%), (55.4%), and (67.7%) respectively.

## VI. CONCLUSION

There are two key benefits to the proposed technique of completely automated binary pattern creation for the detection of finger vein images based on double optimization stages: The first one that in the term of using the deep learning approach we don't need to manually generate our training dataset which improved the lack of the training dataset arability. Second, our approach has also shown a very clear finger vein pattern that is purely shows identical identification cases for the human finger vein image matching. Based on the optimized unsupervised image clustering approach, the proposed system produces a fully unsupervised image clustering and solves the issues that other algorithms such as k-means, FCM, Out's thresholding are facing such as choosing the decent number of clustering, cluster initialization, and automated desired cluster selection. Finally, the proposed system shows a higher accuracy (99.6%) that was achieved by our optimized algorithm, while the second highest score is (76.5), which was achieved by Out's thresholding algorithm.